%% file: iclr2021_conference.tex
\documentclass{article} 
\usepackage{iclr2021_conference,times}
\iclrfinalcopy
\input{math_commands.tex}

\usepackage{hyperref}
\usepackage{url}
\usepackage{graphicx}

\title{Generative modeling of spatio-temporal weather patterns with extreme event conditioning}

\author{Konstantin Klemmer \\
University of Warwick \& TUM \\
\texttt{k.klemmer@warwick.ac.uk} \\
\And
Sudipan Saha \\
TUM \\
\And
Matthias Kahl \\
TUM \\
\And
Tianlin Xu \\
LSE \\
\AND
Xiao Xiang Zhu \\
TUM \& DLR \\
}

\begin{document}

\maketitle

\begin{abstract}
Deep generative models are increasingly used to gain insights in the geospatial data domain, e.g., for climate data. However, most existing approaches work with temporal snapshots or assume $1D$ time-series; few are able to capture spatio-temporal processes simultaneously. Beyond this, Earth-systems data often exhibit highly irregular and complex patterns, for example caused by extreme weather events. Because of climate change, these phenomena are only increasing in frequency. Here, we proposed a novel GAN-based approach for generating spatio-temporal weather patterns conditioned on detected extreme events. Our approach augments GAN generator and discriminator with an encoded extreme weather event segmentation mask. These segmentation masks can be created from raw input using existing event detection frameworks. As such, our approach is highly modular and can be combined with custom GAN architectures. We highlight the applicability of our proposed approach in experiments with real-world surface radiation and zonal wind data.
\end{abstract}

\section{Introduction}
\label{secIntro}
Spatio-temporal modeling is important to many fields, including remote sensing \citep{saha2020change} and climate sciences \citep{racah2017extremeweather}. With the emergence of the big-data era, large scale spatio-temporal datasets have increasingly become available across various domains: from satellite images \citep{sumbul2019bigearthnet} to large-scale mapping efforts \citep{Weber2008}, and also weather data \citep{racah2017extremeweather}, the application area we are looking at in this study. The topic of climate change has gained special attention of the machine learning community in the last few years \citep{rolnick2019tackling, scher2018toward}, while it has been acknowledged that deep learning may provide powerful tools for Earth-systems modeling \citep{Reichstein2019, CampsValls21wiley}. Simulation can be a useful approach for forecasting the future impacts of climate change, allowing decision makers to design data-driven intervention strategies and build a better understanding of how climate change is affecting the planet.
\par
Apart from the great importance of the applications related to spatio-temporal climate and weather data, such data is also very interesting for machine learning researchers as it poses a number of unique challenges: the data is generally multi-modal with distribution shifts and spatio-temporal dependencies. The interactions between different climate and weather indicators is not fully understood \citep{racah2017extremeweather}. Lastly, the data exhibits various anomalies--extreme weather events--a phenomenon only accelerated by climate change.
\par
Although spatio-temporal climate and weather modeling has been studied in past \citep{racah2017extremeweather}, most of the existing work relies on some form of feature engineering. Moreover, studies including temporal analysis often treat the data as 1-D time series and ignore the spatial complexity of the data. Inspired by the success of different generative temporal modeling methods \citep{xiong2018learning, xu2020cot}, in this work we train implicit generative adversarial net (GAN) models optimized for producing spatio-temporal weather patterns. In particular, we examine how prior knowledge on extreme weather events can be incorporated in these generative models. Extreme events can be detected in existing weather data using toolkits like TECA \citep{Prabhat2012}, allowing for the creation of segmentation mask from any given weather data input. As such, the contribution of this study lies in a novel approach for embedding multi-class, spatio-temporal segmentation masks as context vector for conditional GANs. Our proposed method uses a flexible mask encoder, allowing for an easy integration into existing video GANs. We highlight this applicability in experiments with real-world weather data.  
\par
The rest of this paper is organized as follows: we briefly discuss the existing literature in Section \ref{sectionRelatedWork}. Our proposed method is introduced in Section \ref{sectionProposedMethod}.
We present the dataset and our experimental results in Section \ref{sectionExperimentalResult}. We conclude with a brief discussion on potential future work in Section \ref{sectionConclusion}.

\section{Related work}
\label{sectionRelatedWork}
Considering the relevance to our work, here we briefly discuss recent developments related to: \textit{i)} machine learning (ML) based climate and weather modeling and \textit{ii)} spatio-temporal generative adversarial networks.

\subsection{ML based climate and weather modeling}
\label{subsectionMlClimateModeling}
Over the last years, various studies have applied machine learning, especially deep learning, to devise better methods for climate modeling and related challenges. \cite{racah2017extremeweather} presented a multi-channel spatio-temporal CNN architecture that can exploit temporal information and unlabeled climate data to improve the localization of extreme weather events. \cite{scher2018toward} presented a method for forecasting the temporal evolution of weather data. Different deep learning frameworks, e.g., autoencoder \citep{hossain2015forecasting}, CNN \citep{qiu2017short}, and long short-term memory (LSTM) networks \citep{karevan2018spatio} have been used for weather prediction. In \citep{hossain2015forecasting}, an approach for estimating air temperature from historical pressure, humidity, and temperature data is proposed using stacked denoising autoencoder. \cite{qiu2017short} levarage the correlation between multiple sites for weather prediction via  a multi-task CNN. \cite{karevan2018spatio}, use a spatio-temporal LSTM model for temperature prediction of 5 cities in west Europe. \cite{kadow2020artificial} performed monthly reconstructions of the missing climate data via image inpainting and transfer learning. Our proposed model can generate synthetic weather patterns conditioned on extreme event masks and can be useful for supplementing climate data in some of the above-mentioned tasks, e.g. localization of extreme weather events \cite{racah2017extremeweather} and forecasting temporal evolution \cite{scher2018toward}.

\subsection{Spatio-temporal GANs}
\label{subsectionSpatiotemporalGenerativeModeling}
Aligned with its popularity in image modeling, GANs \citep{Goodfellow2014} have been applied to spatial and spatio-temporal modeling, especially video synthesis and analysis: \cite{xiong2018learning} proposed an approach to generate long-term future frames given first frame of a video.
\cite{chen2018lip} devise a model to generate videos from the starting frame and a corresponding audio file. \cite{clark2019adversarial} focus on realistic temporal sequence generation through a dual strategy, employing a spatial discriminator and a temporal discriminator in combination. Several studies have investigated video to video generation / translation, e.g. for video-realistic animations of real humans \cite{liu2019neural}. Beyond video related applications, GANs have also been applied to other spatio-temporal data, e.g. geostatistical data \citep{zheng2020physics} and multi-temporal remote sensing data \citep{saha2019unsupervisedMultisensor, ebel2020}. GANs are further becoming increasingly popular in the GIS community \citep{klemmer2020sxl,Zhu2019}. 
\par
We propose an approach to condition a given GAN with an embedding of the extreme event mask corresponding to the input data. As such, our method can be used in combination with any of the spatio-temporal GANs mentioned here.  

\section{Extreme event conditional GAN}
\label{sectionProposedMethod}

\begin{figure}
    \centering
    \includegraphics[scale=0.8]{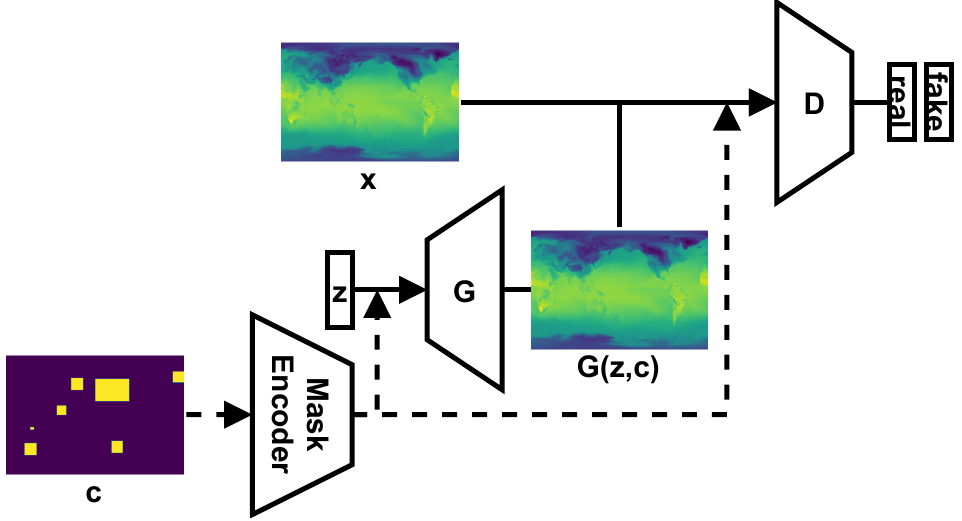}
    \caption{Proposed conditional GAN (agnostic of the customizable GAN backbone) with segmentation mask context, captured through the proposed mask encoder.}
    \label{fig:1}
\end{figure}

Typically, GANs consists of a generator $G$ and a discriminator $D$. $G$ aims to generate realistic synthetic data $G(z)$ from a given random noise input $z$, while $D$ learns to distinguish real data $x$ from fake data $G(z)$. Both networks are then trained by playing a min-max game against each other.
\par
GANs can be conditioned on some context $c$ relevant to the data, such as class labels \citep{mirza2014conditional}. This is facilitated by incorporating the context into both networks, so that the generator output is given as $G(z,c)$ and the discriminator output is $D([x,G(z)],c)$. However, while integrating $c$ is trivial for binary or multi-class labels which can easily presented in vector form, this task is trickier for (multi-class) segmentation masks. We propose a flexible mask encoder to embed the segmentation mask in the desired shape, customizable with any GAN backbone. Extreme event segmentation masks come as binary values in the same shape as the input, comprising one channel per class of event. 
\par
As we require our encoder to capture both the spatial and temporal dynamics of the extreme weather events, we chose an architecture of three $2d$-convolutional layers, each followed by a batch normalization layer and a leaky ReLu activation, in combination with two LSTM layers, again performing batch normalization between the first and second LSTM layer. The output dimension of the last LSTM layer can then be adapted according to the spatio-temporal GAN backbone. This approach follows common practice in dealing with spatio-temporal and video data, first using convolutional layers to encode spatial dynamics, while using LSTM layers to capture the sequential nature of the data. As such, the learnt embedding is able to propagate this information to both generator and discriminator.
\par
To test out our proposed mask encoder, and the general applicability of our approach, we chose a spatio-temporal GAN recently developed by \cite{xu2020cot}: COT-GAN utilizes the sequential nature of temporal data to deploy optimization based on causal optimal transport (COT). This extension of optimal transport ensures that the transformation from source to target probability mass can always only depend on sequential input up to the current time step $t$. The authors find COT-GAN to be efficient, stable and to perform better than comparable approaches. While recent approaches \citep{clark2019adversarial} in video generation rely on a dual discriminator strategy to disentangle spatial and temporal complexities, COT-GAN on the other hand actively integrates the sequential nature of the data into the loss function. To compute the causal loss, COT-GAN also employs a dual discriminator approach (referred to as $h$ and $M$ in the original paper).
\par
Due to these advantages, we chose to apply COT-GAN as the GAN backbone in our experiments, augmenting its discriminators and generator with our aforementioned mask encoder. Figure 1 outlines the proposed GAN framework.

\section{Experimental Results}
\label{sectionExperimentalResult}
\textit{Data:} We test our conditional GAN using two different weather measures and corresponding extreme event segmentation masks, provided by the \textit{ExtremeWeather} dataset \citep{racah2017extremeweather}. We chose to model radiative surface temperature and zonal wind at 850mbar pressure surface. Our training data comprises the whole year of 2004 with four observations per day. At each time step, we extract the variables of interest as well as the extreme event mask, all provided as a $128 \times 196$ pixel image. The segmentation mask denotes the presence of an extreme event with pixel values of $1$, opposed to $0$ values. The segmentation mask comprises four channels, one for each category of extreme event: tropical depression, tropical cyclone, extratropical cyclone and atmospheric river.
\par
\textit{Experimental setting:} We run experiments generating synthetic samples of the measures described above, conditioned on extreme events. Our GANs are trained for 300 epochs using Tesla K80, Tesla T4, and Tesla P100 GPUs. We use the Adam algorithm with decoupled weight decay \citep{Loshchilov2019} to optimize our model.
\par
\textit{Results:} Our results, highlighted in Figure 2, show that training our model for a relatively short time of 300 epochs already provides impressive results, given the complexity of the data input. These initial findings point to a larger potential of implicit generative models for learning climate and weather processes. We believe  that coupled with a more powerful and specialised GAN backbone, our proposed approach might become useful too for simulating and forecasting weather patterns in the presence of extreme and adverse events.

\begin{figure}
    \centering
    \includegraphics[scale=0.22]{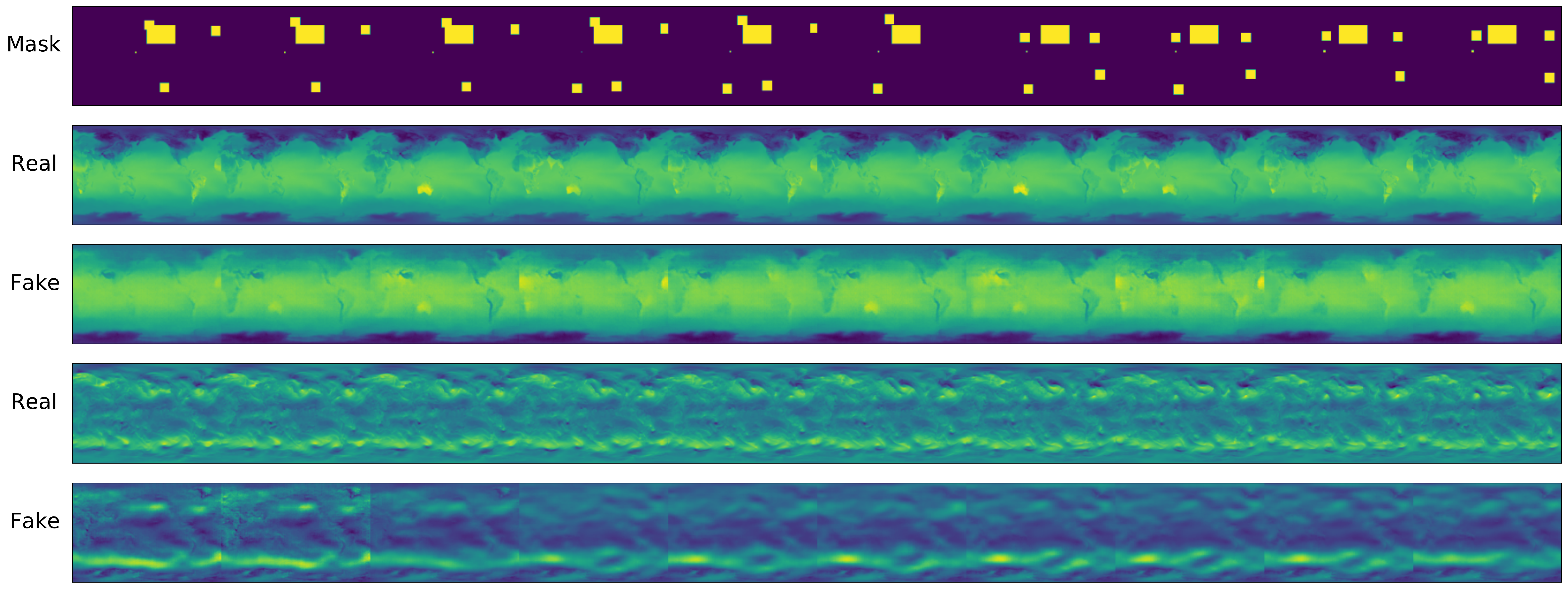}
    \caption{Comparison between real and generated data, including the corresponding extreme event segmentation mask. Row 1 shows a real extreme event segmentation mask for $10$ consecutive time steps. Rows 2-3 show the real and generated radiative surface temperature corresponding to the extreme event pattern. Rows 4-5 show real and generated zonal winds at 850mbar corresponding to the extreme event pattern.}
    \label{fig:2}
\end{figure}

\section{Conclusions}
\label{sectionConclusion}
In this paper, we proposed a conditional GAN based method for generating synthetic spatio-temporal weather patterns. The proposed method effectively ingests extreme event segmentation masks to generate corresponding weather patterns. The generated data shows strong visual
resemblance to real data. The proposed model can be useful for data augmentation in training robust extreme event predictors and weather nowcasting. Our work is an attempt towards better understanding the complex weather patterns and their implication in climate change.
\par
In future studies, we want to build on these preliminary findings. In particular, we want to explore more weather and climate indicators, as well as different extreme events. We further want to devise a more specialized GAN architecture for dealing with this kind of data. Lastly, we seek to explore different options to integrate extreme events into GANs, for example through auxiliary segmentation tasks.

\bibliography{sigproc,iclr_w}
\bibliographystyle{iclr2021_conference}


\end{document}

%% file: math_commands.tex
\usepackage{amsmath,amsfonts,bm}

\def\eqref#1{equation~\ref{#1}}

\def\1{\bm{1}}

\DeclareMathAlphabet{\mathsfit}{\encodingdefault}{\sfdefault}{m}{sl}
\SetMathAlphabet{\mathsfit}{bold}{\encodingdefault}{\sfdefault}{bx}{n}













%% file: iclr2021_conference.bbl
\begin{thebibliography}{26}
\providecommand{\natexlab}[1]{#1}
\providecommand{\url}[1]{\texttt{#1}}
\expandafter\ifx\csname urlstyle\endcsname\relax
  \providecommand{\doi}[1]{doi: #1}\else
  \providecommand{\doi}{doi: \begingroup \urlstyle{rm}\Url}\fi

\bibitem[Camps-Valls et~al.(2021)Camps-Valls, Tuia, Zhu, and
  Reichstein~(Editors)]{CampsValls21wiley}
Gustau Camps-Valls, Devis Tuia, Xiao~Xiang Zhu, and Markus
  Reichstein~(Editors).
\newblock \emph{Deep learning for the Earth Sciences: A comprehensive approach
  to remote sensing, climate science and geosciences}.
\newblock Wiley \& Sons, 2021.
\newblock ISBN 978-1-119-64614-3.

\bibitem[Chen et~al.(2018)Chen, Li, K~Maddox, Duan, and Xu]{chen2018lip}
Lele Chen, Zhiheng Li, Ross K~Maddox, Zhiyao Duan, and Chenliang Xu.
\newblock Lip movements generation at a glance.
\newblock In \emph{Proceedings of the European Conference on Computer Vision
  (ECCV)}, pp.\  520--535, 2018.

\bibitem[Clark et~al.(2019)Clark, Donahue, and Simonyan]{clark2019adversarial}
Aidan Clark, Jeff Donahue, and Karen Simonyan.
\newblock Adversarial video generation on complex datasets.
\newblock \emph{arXiv preprint arXiv:1907.06571}, 2019.

\bibitem[Ebel et~al.(2020)Ebel, Meraner, Schmitt, and Zhu]{ebel2020}
Patrick Ebel, Andrea Meraner, Michael Schmitt, and Xiao~Xiang Zhu.
\newblock Multisensor data fusion for cloud removal in global and all-season
  sentinel-2 imagery.
\newblock \emph{IEEE Transactions on Geoscience and Remote Sensing}, pp.\
  1--13, 2020.
\newblock \doi{10.1109/TGRS.2020.3024744}.

\bibitem[Goodfellow et~al.(2014)Goodfellow, Pouget-Abadie, Mirza, Xu,
  Warde-Farley, Ozair, Courville, and Bengio]{Goodfellow2014}
Ian Goodfellow, Jean Pouget-Abadie, Mehdi Mirza, Bing Xu, David Warde-Farley,
  Sherjil Ozair, Aaron Courville, and Yoshua Bengio.
\newblock {Generative Adversarial Nets}.
\newblock In \emph{Advances in Neural Information Processing Systems
  (NeurIPS)}, pp.\  2672--2680, 2014.
\newblock URL
  \url{http://papers.nips.cc/paper/5423-generative-adversarial-nets}.

\bibitem[Hossain et~al.(2015)Hossain, Rekabdar, Louis, and
  Dascalu]{hossain2015forecasting}
Moinul Hossain, Banafsheh Rekabdar, Sushil~J Louis, and Sergiu Dascalu.
\newblock Forecasting the weather of nevada: A deep learning approach.
\newblock In \emph{2015 international joint conference on neural networks
  (IJCNN)}, pp.\  1--6. IEEE, 2015.

\bibitem[Kadow et~al.(2020)Kadow, Hall, and Ulbrich]{kadow2020artificial}
Christopher Kadow, David~Matthew Hall, and Uwe Ulbrich.
\newblock Artificial intelligence reconstructs missing climate information.
\newblock \emph{Nature Geoscience}, pp.\  1--6, 2020.

\bibitem[Karevan \& Suykens(2018)Karevan and Suykens]{karevan2018spatio}
Zahra Karevan and Johan~AK Suykens.
\newblock Spatio-temporal stacked lstm for temperature prediction in weather
  forecasting.
\newblock \emph{arXiv preprint arXiv:1811.06341}, 2018.

\bibitem[Klemmer \& {B Neill}(2020)Klemmer and {B Neill}]{klemmer2020sxl}
Konstantin Klemmer and Daniel {B Neill}.
\newblock Sxl: Spatially explicit learning of geographic processes with
  auxiliary tasks.
\newblock \emph{arXiv preprint arXiv:2006.08571}, 2020.

\bibitem[Liu et~al.(2019)Liu, Xu, Zollhoefer, Kim, Bernard, Habermann, Wang,
  and Theobalt]{liu2019neural}
Lingjie Liu, Weipeng Xu, Michael Zollhoefer, Hyeongwoo Kim, Florian Bernard,
  Marc Habermann, Wenping Wang, and Christian Theobalt.
\newblock Neural rendering and reenactment of human actor videos.
\newblock \emph{ACM Transactions on Graphics (TOG)}, 38\penalty0 (5):\penalty0
  1--14, 2019.

\bibitem[Loshchilov \& Hutter(2019)Loshchilov and Hutter]{Loshchilov2019}
Ilya Loshchilov and Frank Hutter.
\newblock {Decoupled weight decay regularization}.
\newblock In \emph{7th International Conference on Learning Representations,
  ICLR 2019}, 2019.

\bibitem[Mirza \& Osindero(2014)Mirza and Osindero]{mirza2014conditional}
Mehdi Mirza and Simon Osindero.
\newblock Conditional generative adversarial nets, 2014.

\bibitem[Prabhat et~al.(2012)Prabhat, R{\"{u}}bel, Byna, Wu, Li, Wehner, and
  Bethel]{Prabhat2012}
Prabhat, Oliver R{\"{u}}bel, Surendra Byna, Kesheng Wu, Fuyu Li, Michael
  Wehner, and Wes Bethel.
\newblock {TECA: A parallel toolkit for extreme climate analysis}.
\newblock In \emph{Procedia Computer Science}, 2012.
\newblock \doi{10.1016/j.procs.2012.04.093}.

\bibitem[Qiu et~al.(2017)Qiu, Zhao, Zhang, Huang, Shi, Wang, and
  Chu]{qiu2017short}
Minghui Qiu, Peilin Zhao, Ke~Zhang, Jun Huang, Xing Shi, Xiaoguang Wang, and
  Wei Chu.
\newblock A short-term rainfall prediction model using multi-task convolutional
  neural networks.
\newblock In \emph{2017 IEEE International Conference on Data Mining (ICDM)},
  pp.\  395--404. IEEE, 2017.

\bibitem[Racah et~al.(2017)Racah, Beckham, Maharaj, Kahou, Prabhat, and
  Pal]{racah2017extremeweather}
Evan Racah, Christopher Beckham, Tegan Maharaj, Samira~Ebrahimi Kahou,
  Mr~Prabhat, and Chris Pal.
\newblock Extremeweather: A large-scale climate dataset for semi-supervised
  detection, localization, and understanding of extreme weather events.
\newblock In \emph{Advances in Neural Information Processing Systems}, pp.\
  3402--3413, 2017.

\bibitem[Reichstein et~al.(2019)Reichstein, Camps-Valls, Stevens, Jung,
  Denzler, Carvalhais, and Prabhat]{Reichstein2019}
Markus Reichstein, Gustau Camps-Valls, Bjorn Stevens, Martin Jung, Joachim
  Denzler, Nuno Carvalhais, and Prabhat.
\newblock {Deep learning and process understanding for data-driven Earth system
  science}.
\newblock \emph{Nature}, 566\penalty0 (7743):\penalty0 195--204, feb 2019.
\newblock ISSN 14764687.
\newblock \doi{10.1038/s41586-019-0912-1}.
\newblock URL \url{http://www.nature.com/articles/s41586-019-0912-1}.

\bibitem[Rolnick et~al.(2019)Rolnick, Donti, Kaack, Kochanski, Lacoste,
  Sankaran, Ross, Milojevic-Dupont, Jaques, Waldman-Brown,
  et~al.]{rolnick2019tackling}
David Rolnick, Priya~L Donti, Lynn~H Kaack, Kelly Kochanski, Alexandre Lacoste,
  Kris Sankaran, Andrew~Slavin Ross, Nikola Milojevic-Dupont, Natasha Jaques,
  Anna Waldman-Brown, et~al.
\newblock Tackling climate change with machine learning.
\newblock \emph{arXiv preprint arXiv:1906.05433}, 2019.

\bibitem[Saha et~al.(2019)Saha, Bovolo, and
  Bruzzone]{saha2019unsupervisedMultisensor}
Sudipan Saha, Francesca Bovolo, and Lorenzo Bruzzone.
\newblock Unsupervised multiple-change detection in vhr multisensor images via
  deep-learning based adaptation.
\newblock In \emph{IGARSS 2019-2019 IEEE International Geoscience and Remote
  Sensing Symposium}, pp.\  5033--5036. IEEE, 2019.

\bibitem[Saha et~al.(2020)Saha, Bovolo, and Bruzzone]{saha2020change}
Sudipan Saha, Francesca Bovolo, and Lorenzo Bruzzone.
\newblock Change detection in image time-series using unsupervised lstm.
\newblock \emph{IEEE Geoscience and Remote Sensing Letters}, 2020.

\bibitem[Scher(2018)]{scher2018toward}
Sebastian Scher.
\newblock Toward data-driven weather and climate forecasting: Approximating a
  simple general circulation model with deep learning.
\newblock \emph{Geophysical Research Letters}, 45\penalty0 (22):\penalty0
  12--616, 2018.

\bibitem[Sumbul et~al.(2019)Sumbul, Charfuelan, Demir, and
  Markl]{sumbul2019bigearthnet}
Gencer Sumbul, Marcela Charfuelan, Beg{\"u}m Demir, and Volker Markl.
\newblock Bigearthnet: A large-scale benchmark archive for remote sensing image
  understanding.
\newblock In \emph{IGARSS 2019-2019 IEEE International Geoscience and Remote
  Sensing Symposium}, pp.\  5901--5904. IEEE, 2019.

\bibitem[Weber \& Haklay(2008)Weber and Haklay]{Weber2008}
Patrick Weber and Muki Haklay.
\newblock {OpenStreetMap: user-generated street maps}.
\newblock \emph{IEEE Pervasive Computing}, 2008.

\bibitem[Xiong et~al.(2018)Xiong, Luo, Ma, Liu, and Luo]{xiong2018learning}
Wei Xiong, Wenhan Luo, Lin Ma, Wei Liu, and Jiebo Luo.
\newblock Learning to generate time-lapse videos using multi-stage dynamic
  generative adversarial networks.
\newblock In \emph{Proceedings of the IEEE Conference on Computer Vision and
  Pattern Recognition}, pp.\  2364--2373, 2018.

\bibitem[Xu et~al.(2020)Xu, Wenliang, Munn, and Acciaio]{xu2020cot}
Tianlin Xu, Li~K Wenliang, Michael Munn, and Beatrice Acciaio.
\newblock Cot-gan: Generating sequential data via causal optimal transport.
\newblock In \emph{Advances in Neural Information Processing Systems}, pp.\
  8798--8809, 2020.

\bibitem[Zheng et~al.(2020)Zheng, Zeng, and Karniadakis]{zheng2020physics}
Qiang Zheng, Lingzao Zeng, and George~Em Karniadakis.
\newblock Physics-informed semantic inpainting: Application to geostatistical
  modeling.
\newblock \emph{Journal of Computational Physics}, 419:\penalty0 109676, 2020.

\bibitem[Zhu et~al.(2019)Zhu, Cheng, Zhang, Yao, Gao, and Liu]{Zhu2019}
Di~Zhu, Ximeng Cheng, Fan Zhang, Xin Yao, Yong Gao, and Yu~Liu.
\newblock {Spatial interpolation using conditional generative adversarial
  neural networks}.
\newblock \emph{International Journal of Geographical Information Science},
  pp.\  1--24, apr 2019.
\newblock ISSN 1365-8816.
\newblock \doi{10.1080/1365881YYxxxxxxxx}.
\newblock URL
  \url{https://www.tandfonline.com/doi/full/10.1080/13658816.2019.1599122}.

\end{thebibliography}
